\pdfoutput=1

\documentclass[11pt]{article}

\usepackage{coling}

\usepackage{times}
\usepackage{latexsym}

\usepackage[T1]{fontenc}

\usepackage[utf8]{inputenc}
\usepackage{algorithm}
\usepackage{algpseudocode}
\usepackage{multirow}
\usepackage{enumitem}
\usepackage{csquotes}

\usepackage{tikz}
\usetikzlibrary{shapes.geometric, arrows, calc, decorations.pathreplacing, positioning}
\usepackage{varwidth}
\tikzstyle{startstop} = [rectangle, rounded corners, 
minimum width=2cm, 
minimum height=1cm,
text centered, 
draw=black, 
fill=red!30]

\tikzstyle{io} = [trapezium, 
trapezium stretches=true, 
trapezium left angle=70, 
trapezium right angle=110, 
minimum width=2cm, 
minimum height=1cm,
text width=2.6cm, 
text centered, 
draw=black, fill=blue!30]

\tikzstyle{process} = [rectangle, 
minimum width=2.7cm, 
minimum height=1cm,
text centered, 
text width=2.7cm, 
draw=black, 
fill=orange!30]

\tikzstyle{decision} = [diamond, 
minimum width=2cm, 
minimum height=1cm, 
text centered, 
draw=black, 
fill=green!30]

\newcommand{\arsp}{2mm}
\tikzset{
    twoway/.style={
        decoration={
            show path construction,
            lineto code={
                \path (\tikzinputsegmentfirst); \pgfgetlastxy{\a}{\b};
                \path (\tikzinputsegmentlast); \pgfgetlastxy{\x}{\y};
                \coordinate (uu) at ($(0,0)!.5*\arsp!(\b-\y,\x-\a)$);
                \draw[-latex]($(\a,\b)+(uu)$) to ($(\x,\y)+(uu)$);
                \draw[latex-]($(\a,\b)-(uu)$) to ($(\x,\y)-(uu)$);
            }
        }, decorate
    } 
}

\usepackage{microtype}

\usepackage{inconsolata}

\usepackage{enumitem}
\usepackage{authblk}

\title{Contextual ASR Error Handling with LLMs Augmentation for Goal-Oriented Conversational AI}

\author[1]{Yuya Asano}
\author[1]{Sabit Hassan}
\author[1]{Paras Sharma}
\author[2]{Anthony Sicilia}
\author[2]{Katherine Atwell}
\author[1]{\\Diane Litman}
\author[2]{Malihe Alikhani}

\affil[1]{University of Pittsburgh}
\affil[2]{Northeastern University}
\affil[ ]{\texttt{\{yua17, sabit.hassan, pas252, dlitman\}@pitt.edu}}
\affil[ ]{\texttt{\{sicilia.a, atwell.ka, m.alikhani\}@northeastern.edu}}

\begin{document}
\maketitle
\begin{abstract}
General-purpose automatic speech recognition (ASR) systems do not always perform well in goal-oriented dialogue. Existing ASR correction methods rely on prior user data or named entities. We extend correction to tasks that have no prior user data and exhibit linguistic flexibility such as lexical and syntactic variations. We propose a novel context augmentation with a large language model and a ranking strategy that incorporates contextual information from the dialogue states of a goal-oriented conversational AI and its tasks. Our method ranks (1) $n$-best ASR hypotheses by their lexical and semantic similarity with context and (2) context by phonetic correspondence with ASR hypotheses. Evaluated in home improvement and cooking domains with real-world users, our method improves recall and F1 of correction by 34\% and 16\%, respectively, while maintaining precision and false positive rate. Users rated .8-1 point (out of 5) higher when our correction method worked properly, with no decrease due to false positives.
\end{abstract}

\section{Introduction and Related Work}\label{introduction}


Although domain-agnostic automatic speech recognition (ASR) models are improving, advanced models still make errors, hindering fluent dialogues with conversational AI \citep{pmlr-v202-radford23a}. 
Most prior work on context-based ASR error correction needs prior user interaction data to model errors statistically \citep{sarma-palmer-2004-context,jonson2006dialogue,shivakumar2019learning,liu2019semantic,ponnusamy2022feedback,lopez2008asr,weng2020joint,zhou-etal-2023-unified}, make natural language understanding robust to errors \citep{gupta-etal-2019-casa}, or learn user preferences \citep{raghuvanshi-etal-2019-entity,cho-etal-2021-personalized,bis-etal-2022-paige}. These data are not always available, especially when creating AI in a new domain. 

One proposed solution is to use tasks that goal-oriented conversational AI can accomplish as a primary source of context. It includes restricting ASR hypotheses to the vocabularies that a natural language understanding module can parse \citep{he2003data,whittaker1995advanced} and remembering named entities and retrieving the most probable one based on textual and phonetic similarities and $n$-best ASR hypotheses \citep{georgila2003speech,raghuvanshi-etal-2019-entity,wang21b_interspeech,bekal2021remember}. However, it overlooks three types of flexibility in natural language in Table \ref{tab:variations}: lexical and syntactic variations for the same goal, non-essential modifiers inserted in the middle of task phrases, and the omission of words that are not needed to specify a goal given a context. An ASR system could also omit some words in error.

\begin{figure}[t]
    \centering
    \includegraphics[width=0.45\textwidth]{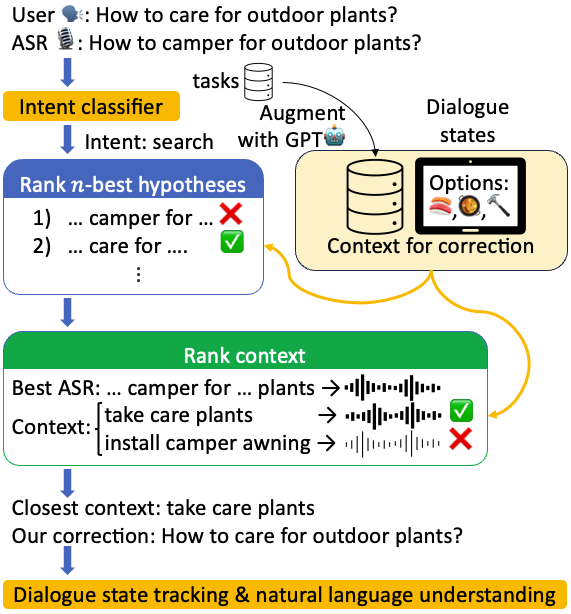}
    \vspace{-0.08in}
    \caption{The overview of our ASR correction pipeline. After intent classification, we re-rank $n$-best ASR hypotheses by the lexical and semantic similarity of context retrieved from the dialogue states and the GPT-augmented task list. If none of the hypotheses seems plausible, we find a task or option that sounds phonetically similar to the best hypothesis.}
    \label{fig:flowchart}
\end{figure}





\begin{table*}[t]
    \begin{tabular}{p{2.5cm}|p{3cm}|p{8.4cm}}
        Types & Original & Variations \\
        \hline
        Lexical \& syntactic variations & How can I make a wood fence? & How can a wood fence be built? \\
        \hline
        Inserting modifiers & Make a wood fence & Make a wood privacy fence (this should not be mapped to ``make privacy in a college dorm.'')\\
        \hline
        Omitting unneeded words & I want to build a wood fence. & I want to build a fence. (options are ``build a wood fence,'' ``paint a wood fence,'' and ``clean a wood fence.'')
    \end{tabular}
    \vspace{-0.06in}
    \caption{Examples of the variations of task queries. These examples and all other examples in this paper are not real-user conversations but resemble them.}
    \label{tab:variations}
\end{table*}

To tackle these challenges, we extend the work on the use of tasks as context with
\setlist{nolistsep}
\begin{enumerate}[noitemsep]
    \item better ranking strategies robust to insertions and omissions of tokens,
    \item reduction of the size of the context using the dialogue state and augmentation with partial matches (for token omissions), and
    \item offline augmentation of tasks with a large language model (LLM) (for lexical and syntactic varieties).
\end{enumerate}
These ranking and augmentation strategies are system-agnostic and generalizable to any tasks if a list of the supported tasks is available. They are lightweight and are not affected by the latency of LLMs. The overview of our method is in Figure \ref{fig:flowchart}. 

We deploy our method on Amazon Alexa and test its effectiveness with real-world users. It improved recall and F1 of correction from a baseline while keeping fair precision and false positive rate (FPR) in offline evaluation. In online evaluation, our ratings have increased by .8-1 point out of 5 when our method corrected errors properly and did not decrease even with false positives.

\section{Problem Description and Data}\label{dataset}

\begin{figure*}[t]
\centering
\includegraphics[width=0.9\textwidth]{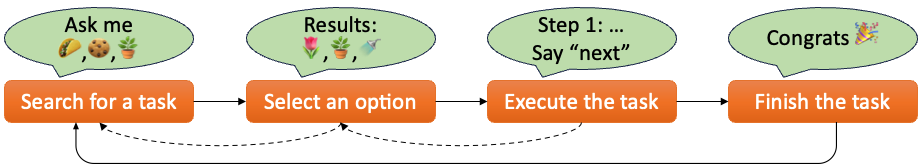}
\vspace{-0.08in}
\caption{The flow of dialogues in our dataset. The most probable flows are in solid arrows. A dashed arrow indicates that the path is possible but unlikely compared to the solid arrows.}
\label{fig:dialogue-flow}
\end{figure*}





We define the goal of ASR correction for goal-oriented conversational AI as to \textit{pass the corrected information from erroneous ASR transcripts to the following modules} (in our case, dialogue state tracking and natural language understanding in Figure \ref{fig:flowchart}) \textit{so that AI can take the correct action as requested by a user}. This definition is similar to call routing, which ``is concerned with determining a caller’s task'' \cite{williams2004comparison}.

We implemented our system through Alexa Skills,\footnote{\url{https://developer.amazon.com/en-US/alexa/alexa-skills-kit}} which allows third-party developers to create conversational AI on Alexa. Our Alexa Skill assists with cooking and home improvement tasks.
After our system's introduction (cf. Appendix \ref{preamble}), users typically start with searching for a task (e.g., ``How to fix a faucet?'') as illustrated in Figure \ref{fig:dialogue-flow}. The system replies with search results from an internal dataset of Whole Foods recipes and WikiHow articles. Users select an option and complete tasks step-by-step using voice commands (e.g., ``next,'' ``go back''), instructions from our system (e.g., our system says, ``Say `start cooking' when you're ready for the recipe steps.''), or Alexa's touch screen. An example dialogue can be found in Appendix \ref{example}. Alexa's ASR system provides up to 5-best hypotheses.

For evaluation, we sampled thousands of dialogues between March 28, 2023, and August 5, 2023, as part of the Alexa Prize TaskBot Challenge 2 \citep{Agichtein2023}. One author manually annotated ASR errors and provided correct transcripts. 4.1\% of the utterances had ASR errors and 18.2\% of dialogues experienced at least one error. Of these, 51\% happened during search (the left most dialogue state in Figure \ref{fig:dialogue-flow}), 17\% during selection (the second left state in Figure \ref{fig:dialogue-flow}), and 13\% during task execution (the second right state in Figure \ref{fig:dialogue-flow}).\footnote{The numbers do not add up to 100\% because some errors happened when intents fell into none of these three intents.}

\section{Proposed Method}
Our method starts with identifying likely user responses that can be used as context for correction and deciding when to trigger the correction based on dialogue states (Section \ref{state}). Next, we re-rank $n$-best ASR hypotheses using this context (Section \ref{nbest}). If none is plausible, we rank the context using phonetic information to fix errors in the best ASR hypothesis (Section \ref{phoneme}). To handle lexical and syntactic varieties and word omissions, we augment the context and task list for correction (Section \ref{augmentation}). An overview of our pipeline is in Figure \ref{fig:flowchart}.

\subsection{Dialogue states as context}\label{state}

The first step of our method is identifying a context to correct ASR errors. Dialogue states help predict what a user might say next. 
For example, users often select an option after seeing search results (cf. Figure \ref{fig:dialogue-flow}) or pick a system-suggested query at the start (cf. Appendix \ref{preamble}). Therefore, we define a \textit{narrow context} as a list of likely user responses based on the current dialogue state. Our ranking method matches a user’s utterance to this context. In our case, a narrow context is the options presented by the system in the previous turn or the voice commands to execute tasks.

However, users may not always respond within the scope of a narrow context, making it difficult for the ranking methods in sections \ref{nbest} and \ref{phoneme} to find matches, especially when starting a new task. In this case, we retrieve context from all tasks using an indexed search for the ASR hypotheses to get lexically overlapping tasks. We measure the cosine similarity between the hypotheses and the search results using Sentence-BERT \citep{reimers-gurevych-2019-sentence} embeddings and retain only highly similar results (see sections \ref{nbest} and \ref{phoneme} for the actual thresholds) as context. Then, we rerun the methods with the new context.

To reduce false positives, our correction is triggered only in specific dialogue states. It activates when a narrow context exists (the two middle states in Figure \ref{fig:dialogue-flow} and at the beginning of a conversation) or the system's intent classifier predicts that the user intends to search for or select a task and such intent is probable in the current state (the two left states in Figure \ref{fig:dialogue-flow}). Our method is not triggered when, for example, the user tries to ask a question about the task they are executing, or the system asks a question to them. Appendix \ref{state-tracking} provides details on deriving a narrow context and determining when to trigger the method.

\subsection{Re-ranking $n$-best ASR hypotheses}\label{nbest}
We re-rank ASR hypotheses by the lexical and semantic similarity to the context. If a narrow context exists, we score hypotheses from best to worst by fuzzy matching \citep{chaudhuri2003robust} with the context. We stop if we find one exceeding a threshold.\footnote{We set the threshold to 96\% to allow small mistakes such as articles (``a'' vs ``the'') and singular vs plural.} If no match is found, we run an indexed search\footnote{We tuned the threshold for the indexed search using some examples from the post-hoc analysis and set it to 0.8.} for each ASR hypothesis and assign the largest cosine similarity as the score of the hypothesis. 
The highest-scored hypothesis is chosen as a correction. If the original best hypothesis is picked, we interpret this as no correction needed.

Re-ranking $n$-best hypotheses handles modifier insertions. Suppose ``how to care for outdoor plants'' is transcribed as ``how to camper for outdoor plants'' and that $n$-best hypotheses have the correct transcription. ``How to camper for outdoor plants'' does not return a good search result since there is no such task. However, even if ``how to care for outdoor plants'' is not on the list, our method can propose it as a correction because its search result, ``take care plant,'' exceeds the threshold.

\subsection{Ranking context}\label{phoneme}

Since $n$-best ASR hypotheses don’t always include the correct transcripts, re-ranking them can fail. We solve this by ranking context based on phonetic similarity to the best ASR hypothesis measured by the longest common subsequences (LCSs) between the phoneme sequences of the context and the best ASR hypothesis\footnote{Phoneme sequences are obtained from g2pE \citep{g2pE2019}, which looks up the CMU Pronouncing Dictionary \citep{lenzo2007cmu} or predicts the pronunciation using a neural network model if a word is not in the dictionary.}. 
We choose the option from the context whose LCS covers a certain portion of the option and is not too scattered as a candidate for correction. Then, we replace the tokens in the best hypothesis covered by the LCS with those in the candidate and accept the new hypothesis as the correction. The full algorithm is in Appendix \ref{context-code}.

LCSs help handle phrases inserted or removed in the middle of a task. Suppose there is a task called ``how to fix a bathroom faucet'' and that ASR transcribes ``how can I fix a leaky bathroom faucet'' as ``how can I fix a leaky bathroom for sit.'' The LCS between the best ASR hypothesis and the context comes from ``fix a bathroom for sit'' for the hypothesis. Therefore, we can skip the word ``leaky'' in the original hypothesis to replace ``for sit'' with ``faucet'' from the context to get the correct transcript. The same mechanism works when there is a task called ``how to fix a leaky bathroom faucet'' and ASR transcribes ``how can I fix a bathroom faucet'' as ``how can I fix a bathroom for sit.''

\begin{table*}[t]
    \centering
    \begin{tabular}{c|cccc|cccc|cccc}
        \hline
        Intent & \multicolumn{4}{c|}{Search (91.6\%)} & \multicolumn{4}{c|}{Selection (8.4\%)} & \multicolumn{4}{c}{Combined} \\
         & Prec & Rec & F1 & FPR & Prec & Rec & F1 & FPR & Prec & Rec & F1 & FPR \\
        \hline
        Baseline & .17 & .04 & .06 & \textbf{.01} & .92 & .47 & .62 & \textbf{.01} & \textbf{.56} & .18 & .27 & \textbf{.01} \\
        Baseline + GPT & .14 & .03 & .05 & \textbf{.01} & .92 & .47 & .62 & \textbf{.01} & .55 & .17 & .26 & \textbf{.01} \\
        \hline
        Ours & \textbf{.22} & \textbf{.38} & \textbf{.28} & .05 & \textbf{.98} & \textbf{.86} & \textbf{.91} & \textbf{.01} & .37 & \textbf{.54} & \textbf{.44} & .04 \\
        \hline
    \end{tabular}
    \vspace{-0.08in}
    \caption{The precision (Prec), recall (Rec), F1 score, and false-positive rate (FPR) of corrections from the baseline (cf. section \ref{baseline}) and our method, broken down by search and selection intents, as well as both intents combined. The best results are in bold. Our method outperformed the baseline in both intents, except for FPR in search. The baseline has the highest precision for the two intents combined due to low recall and FPR in the search intent.}
    \label{tab:result}
\end{table*}

\begin{table*}[t]
    \centering
    \begin{tabular}{p{3.5cm}|p{4cm}|p{3.5cm}|p{3.5cm}}
        Original ASR & Correct transcript & Baseline & Ours \\
        \hline
        Alexa how do I choose without polish? & Alexa how do I shine shoes without polish? & Alexa how do I choose without polish? & Alexa how do I shine shoes without polish? \\
        \hline
        Cartoon electric guitar & How to tune electric guitar & Cartoon electric guitar & Tune an electric guitar \\
        \hline
        How to make a snowflake of paper & How to make a snowflake out of paper & How to make a snowflake out of paper & How to make a snowflake out of paper \\
        \hline
        Start another task & Start another task & Stay on task while working on a computer & Start another task
    \end{tabular}
    \vspace{-0.08in}
    \caption{The correct and original ASR transcripts and the correction made by the baseline and our method.}
    \label{tab:example-correction}
\end{table*}

\subsection{Augmenting context}\label{augmentation}

Ranking $n$-best ASR hypotheses and context depends on the richness of the context to better deal with erroneous hypotheses and lexical and syntactic variety of user utterances. Therefore, we augment the context used for ranking in two ways.

First, we expand the search space by distilling lexical and syntactic variations from GPT to create an offline dataset, allowing ASR correction to consider phrases not in the list of available tasks. To avoid high costs and repetitive data, we cluster the task list and generate variations only for cluster centroids. This allows for a collection of the most representative yet diverse samples \cite{hassan-alikhani-2023-calm}. Then, we index the augmented dataset for an indexed search. The details of the implementation can be found in Appendix \ref{augment-detail}, and the evaluation is in Appendix \ref{search-examples}. 

Second, we add partial matches to a narrow context if they uniquely identify one option. For example, if the search results are ``how to care for indoor plants,'' ``how to water indoor plants,'' and ``how to fertilize indoor plants,'' we suggest ``how to water indoor plants'' as long as ASR correctly transcribes ``water.'' Partial matches also handle the omission of unneeded words.

\section{Experiments}\label{experiment}

\subsection{Post-hoc analysis for search \& selection}\label{quantitative}
We did a post-hoc analysis on the user utterances to search or select home improvement (wikiHow) tasks because they have more linguistic variations than recipes.
We extracted these utterances from the annotated dataset in section \ref{dataset}. 91.6\% of this subset had the search intent. Since the goal of our ASR correction is to identify the desired intent (cf. Section \ref{dataset}), a proposed correction was considered correct if it matched the manually corrected transcript or the correct option. We evaluated the methods with Precision@1, Recall@1, and F1-score@1 since our method does not rank beyond the first option, and the scores of our two ranking strategies have different meanings.

\subsubsection{Baseline}\label{baseline}
We use the method by \citet{raghuvanshi-etal-2019-entity} as a baseline because it does not require prior user interactions and its source code is publicly available. This method matches potentially erroneous ASR output with named entities and their synonyms stored in a database, based on textual and phonetic similarities aided by $n$-best ASR hypotheses. We treated each wikiHow article title in the private dataset as a named entity. We also experimented with adding alternative titles from GPT (the same as section \ref{augmentation}) as synonyms of named entities. 

\subsubsection{Results}

We examined the search and selection intents separately and combined. The results are summarized in Table \ref{tab:result}. Our method has \textit{36\% higher recall and 17\% higher F1} than the baseline overall. Although its overall precision was 19\% lower and its FPR was 3\% higher, our method \textit{outperformed the baseline in every metric except for FPR in the search intent} when broken down by intent. This is because over 80\% of the utterances our method made corrections had the search intent, while the baseline had fewer than 50\%. Thus, combined precision favored the search intent (harder) for our method and the selection intent (easier) for the baseline. GPT augmentation worsened the baseline's precision and recall for the search intent. 

Table \ref{tab:example-correction} shows examples where our method corrected distorted transcripts (the first two rows), while the baseline only fixed small errors such as particles (the third row). About 93\% of the baseline’s corrections were also made by our method. The last row is a case where the baseline falsely chose an alternative with one or two words overlapping the original transcript.

We also compared the word error rate (WER) in Table \ref{tab:wer}. Our method showed a marginal .5\% increase compared to the original ASR transcript ($p<.1$), but no significant difference from the baseline ($p>.6$) with a t-test and adjusted p-values using the Holm method. This increase does not contradict the higher precision, recall, and F1 scores because about 97\% of the utterances with an increased WER are false positives.

\begin{table}[t]
    \centering
    \begin{tabular}{c|c|c}
        WER & M & SD \\
        \hline
        No correction method & .015 & .085 \\
        Baseline & .021 & .123 \\
        Ours & .020 & .079
    \end{tabular}
    \vspace{-0.08in}
    \caption{The means (M) and standard deviations (SD) of word error rate (WER) with no correction method, the baseline, and ours. An increase in WER is due to false positives, but the overall identification rate of tasks requested by users is vastly improved (cf. Table \ref{tab:result}).}
    \label{tab:wer}
\end{table}

We did ablation studies to assess the contribution of the two ranking strategies. Table \ref{tab:ablation} shows that ranking $n$-best is better for the search intent while ranking context by phonemes is better for the selection intent. Combining the two boosts precision and recall for selection by 6-23\%, as phonemes catch cases missed by $n$-best re-ranking when the hypotheses do not contain the exact match with narrow context. However, combining the two does not improve recall and worsens precision, F1, and FPR for the search intent.

\begin{table*}[t]
    \centering
    \begin{tabular}{c|cccc|cccc|cccc}
        \hline
        Intent & \multicolumn{4}{c|}{Search (91.6\%)} & \multicolumn{4}{c|}{Selection (8.4\%)} & \multicolumn{4}{c}{Combined} \\
         & Prec & Rec & F1 & FPR & Prec & Rec & F1 & FPR & Prec & Rec & F1 & FPR \\
        \hline
        $n$-best + augment & +.07 & .00 & +.05 & -.02 & -.10 & -.23 & -.18 & -.01 & +.04 & -.08 & .00 & -.01 \\
        Phoneme + augment & -.11 & -.15 & -.16 & -.01 & -.06 & -.13 & -.10 & .00 & -.08 & -.21 & -.13 & .00 \\
        \hline
    \end{tabular}
    \vspace{-0.08in}
    \caption{The results of the ablation studies (the numbers are relative to our full method at the bottom of Table \ref{tab:result}).}
    \label{tab:ablation}
\end{table*}

We further performed an error analysis to understand failure cases, with examples in Appendix \ref{examples}.\\
\textbf{Errors our method could not correct} \quad Our method failed for the search intent primarily for five reasons. 41\% of the failures were due to errors in key phrases in all hypotheses. In this case, the correct parts were not sufficient to bring up good search results. 
14\% were due to missing proper nouns not in the augmented dataset. 12\% were due to errors in prepositions not captured by cosine similarity. 
12\% were caused by incorrect choices from $n$-best ASR hypotheses when the correct one did not return a good search result during ranking. 
12\% occurred when incorrect hypotheses scored high during ranking.

 

For the selection intent, our method missed errors only when the correct transcripts were absent or when the errors were large (e.g., ``automatic writing'' misheard as ``ornamented lightning'').\\
\textbf{False positives due to a narrow context} \quad 51\% of false positives were caused by prioritizing a narrow context: 
(1) users requesting to exit our system by opening another app on Alexa (39\%), (2) users making similar queries to be more specific or correct ASR by themselves (36\%), (3) ASR hypotheses including numbers (our method suggested it as correction because our system allows users to select an option by number like ``option three'' (8\%), and (4) users asking for a chit-chat (5\%). Better intent classification could address (1) and (4).\\
\textbf{False positives not caused by a narrow context} \quad Bad index search results gave high scores to incorrect hypotheses (40\%) and ungrammatical hypotheses (9\%) or did not give high scores to any hypotheses (24\%). 
Removing (sub) words was prevalent as well (27\%) but the main intents were mostly preserved.


\subsection{Real-world deployment}\label{real-world}
We incrementally deployed our method for matching the system's suggestions (part of the search), selection, and keywords for task execution for recipes and home improvement with real-world users. To prevent dialogue disruption by false positives, our system asked ``Did you mean <correction>?'' when suggesting a correction. Users rated the quality of the conversations from 1 (worst) to 5 (best). We compared the ratings of users who experienced proper correction by our method (true positive), improper correction (false positive), no ASR errors and no correction (true negative), and some ASR errors not handled by our method (false negative) in Table \ref{tab:rating}.\footnote{A user may experience two or more categories in one interaction. We removed such users from our analysis.} We ran one-way ANCOVA to control for the number of turns and the timestamps of the conversations since we also deployed other new features and bug fixes incrementally. We adjusted p-values with Holm's method. There was a significant difference between true positive and false negative ($p<.05$) but not between false positive and false negative ($p>.1$) or between all positives and all negatives ($p>.1$). In the latter comparisons, the difference was attributed to the number of turns. The results imply we maintained or even improved the score even with false positives.

\begin{table}[t]
    \centering
    \begin{tabular}{c|cc|cc}
        & \multicolumn{2}{|c|}{Turns} & \multicolumn{2}{c}{Rating}\\
        ASR errors & M & SD & M & SD \\
        \hline
        True positive (TP) & 12.6 & 9.32 & 3.98 & 1.40 \\
        False positive (FP) & 8.29 & 2.75 & 2.29 & 1.38 \\
        TP + FP & 11.5 & 8.30 & 3.54 & 1.56 \\
        \hline
        True negative (TN) & 7.70 & 18.6 & 3.13 & 1.58 \\
        False negative (FN) & 13.4 & 17.0 & 2.89 & 1.56 \\
        TN + FN & 8.45 & 10.1 & 3.10 & 1.64 
    \end{tabular}
    \vspace{-0.08in}
    \caption{The mean (M) and standard deviations (SD) of the user turns and rating of the conversations that experienced proper correction (true positive), improper correction (false positive), no ASR errors and no correction (true negative), and ASR errors not corrected (false negative). Note that true and false negatives may receive correction by our full method but did not at the time of the interactions due to incremental deployment.}
    \label{tab:rating}
\end{table}

\section{Discussion and Conclusion}

Our approach corrects ASR errors in general terms in goal-oriented dialogues by leveraging dialogue context more flexibly than existing methods, which focus on named entities. Context is taken from tasks and narrowed down by the dialogue state. We augment a narrow context via partial matches and tasks with GPT to handle linguistic variability. We re-rank $n$-best ASR hypotheses based on the semantic similarity with context and rank context by phonetic correspondence with ASR hypotheses. Unlike existing methods, our method requires no prior user interaction data. Experiments show it improves recall and F1 from the baseline while keeping reasonable precision and FPR. In a real-world deployment, it improved user ratings when correct and had minimal impact when incorrect.

Though evaluated in home improvement and cooking, our approach can be applied to any domain or voice commands, provided there is a comprehensive task list for LLM augmentation. Although it assumes correct intent recognition, it can also enhance intent recognition if textual cues for intents can be extracted from training data. Moreover, although designed for goal-oriented dialogues where user speech is often constrained by their goals, our method can also be adapted to open chat by, for example, generating a narrow context through simulations with an LLM. Future work can evaluate it in more domains and dialogue states.

The drawback of our method is an increase in FPR when correcting errors using all tasks. To reduce it, we could consider phoneme similarities. Suppose ASR transcribed ``house'' as ``horse'' when available options include ``house'' and ``hence.'' g2pE converts ``house'' to [``HH,'' ``AW1,'' ``S''], ``horse'' to [``HH,'' ``AO1,'' ``R,'' ``S''], and ``hence'' to [``HH,'' ``EH1,'' ``N,'' ``S'']. Since LCSs do not consider similarities of phonemes, both ``house'' and ``hence'' have the same length of LCS as ``horse.'' However, intuitively, ``AO1'' should sound closer to ``AW1'' than ``EH1,'' so ``horse'' should be corrected to ``house.'' This could be potentially solved by using phoneme embedding to weigh the differences among phonemes \citep{fang2020using} and solving the heaviest LCS problem \citep{jacobson1992heaviest}. This could be also addressed by an audio tokenizer and the subsequent embedding in a pre-trained speech LLM \citep{borsos2023audiolm}.

\section*{Acknowledgements}
This project was completed as part of and received funding from the Alexa Prize TaskBot Challenge 2. We would like to thank the Alexa Prize team, especially Lavina Vaz and Michael Johnston, for supporting us throughout the competition and for giving us the resources to develop and deploy our system to a large audience. We would also like to thank our team members who are not in the authors list for making our research possible: Mert Inan, Qi Cheng, Dipunj Gupta, and Jennifer Nwogu. We would like to thank members in PETAL lab and FACET lab at the University of Pittsburgh for giving us valuable feedback.

\section*{Limitations}
We acknowledge the limitations of the evaluation of our method. First, while public benchmarks focusing on ASR errors exist, we evaluate our method only on our private data instead because those public benchmarks typically do not contain the broader context of functional conversational AI. We argue that not evaluating our method against public benchmarks does not affect the validity of our approach in practical applications, as the primary objective of ASR error correction lies in enhancing the performance of downstream conversational AI. Although Amazon's policy prevents this paper from reporting some statistics in our dataset (e.g., the number of sampled dialogues), listing real examples (we modify the user examples listed in this paper but preserve actual errors from Alexa's ASR), and releasing our code, we provide the necessary details so that the industry community can easily benchmark our method in various domains, including novel areas that do not have any prior user interaction data, when developing a conversational AI. This paper should serve as a cornerstone of the advancement of ASR correction in the presence of linguistic flexibility in practical real-world conversational AI applications.

Second, we assess our algorithm's performance using precision, recall, and F1-score only at rank one. As our algorithm stops ranking upon identifying a suitable candidate to prevent over-correction, it restricts the calculation of Precision@$N$, Recall@$N$, and F1-score@$N$ for $N >= 2$. In addition, Precision@$N$, Recall@$N$, and F1-score@$N$ will be always 1 for the selection intent and for $N >= 3$ because we present only three options to users. That being said, we intend to update the algorithm by adding multiple candidate considerations and integrating varied selection criteria for candidate options in future iterations.

We are also aware of a more advanced grapheme-to-phoneme conversion model based on a transformer.\footnote{\url{https://github.com/cmusphinx/g2p-seq2seq}} We decided not to use it due to the restriction on latency imposed on Alexa applications. The discussion of the performance and latency of our method compared to existing deep learning approaches such as \citet{bekal2021remember} and \citet{mai2024enhancing} and the online use of LLMs \citep{chen2023hyporadise} is left for future work.

\section*{Ethical Considerations}
ASR is known to be biased against dialects, females, and racial minorities \citep{tatman-2017-gender,tatman2017effects}, possibly due to a lack of their representation in training datasets. ASR struggles with transcribing low-resource languages, too \citep{zellou2024linguistic}. Our work could potentially aid the experience of marginalized populations with conversational AI as it does not rely on any data other than a list of tasks/commands. However, this might not be the case because the effectiveness of our method depends on the quality of ASR, namely $n$-best hypotheses and phonetic information, and augmentation by LLMs, which also have poor performance on low-resource languages \citep{hangya-etal-2022-improving}. A closer investigation of how well our method fills a gap between marginalized populations and white males speaking general English is needed.

\bibliography{anthology,custom}

\appendix

\section{Our system's preamble}\label{preamble}
When a user launches our system, it says something like the following:
\begin{displayquote}
Hi, this is <our system's name>. Welcome! I can help with a task you choose. You can ask me for things like recipes for cookies, how to fix a faucet, or how to make origami.
\end{displayquote}
In this example, the suggestions from the system are ``cookies,'' ``how to fix a faucet,'' and ``how to make origami.''

\section{Example dialogue}\label{example}
Table \ref{tab:example-dialogue} shows an example of a dialogue between a user and our system.
\begin{table}[ht]
    \centering
    \begin{tabular}{c|p{5.8cm}}
        Speaker & Utterance \\
        \hline
        System & Hi, this is <our system's name>. Welcome! I can help with a task you choose. You can ask me for things like recipes for cookies, how to fix a faucet, or how to make origami. \\
        User & How can I make salsa? \\
        System & Let's spice things up in the kitchen with these tasty Whole Foods recipes! Here's what I found for your search on salsa: The top results are Salsa, Pico de Gallo (Fresh Salsa), and Homemade Salsa. Other users have rated the first one pretty high, and for a quick recipe, check out the second one. Don't forget to select an option, summarize it, or search for something else. \\
        User & Fresh salsa. \\
        System & Great! Before we get started, please be careful when using any tools or equipment. Remember, safety first! OK, here are the ingredients for Pico de Gallo (Fresh Salsa). The first ingredient is: 2 cups chopped tomatoes. Say ``next'' when you're ready for another. Or say ``start cooking'' when you're ready for the recipe steps. \\
        User & Start cooking. \\
        System & Okay, here's the first step. Combine all ingredients in a bowl, toss well and serve chilled or at room temperature. \\
        User & Next. \\
        System & There are no more steps left. Would you like to finish this recipe? \\
        User & Yes. \\
        System & Congratulations! We completed the task! If you want help with another task, just ask me. \\
        User & Alexa, cancel. \\
        & (End of conversation)
    \end{tabular}
    \caption{An example of an interaction between a user and our system.}
    \label{tab:example-dialogue}
\end{table}

\section{Details of our dialogue state tracking}\label{state-tracking}
To track the current dialogue state, we store numerous values from the current and previous turn(s). These include whether a question is just asked by the system, whether there exists an unanswered question from the previous turn, search results (if there are any) from the previous turn, the task (if any) the user is currently working on, and values extracted from our natural language understanding module indicating the user's intent and dialogue slots for the current turn. These variables allow the system to, at any given turn, track the current dialogue stage (cf. Figure \ref{fig:dialogue-flow}) and access any relevant contextual information. 

We also implemented intent recognition and slot filling enhanced by the rich contextual information coming from our state tracking. For example, it was used to interpret the results of lightweight classifiers (e.g., rule-based classifiers for the intent to end the conversation and for fine-grained intents that occur at specific points within a task, such as the intent to go to the next step of a task). Database-specific information, including frequently used verbs in our datasets, was collected and used to further inform our slot-filling algorithm. 

\section{Pseudocode for ranking context}\label{context-code}

\begin{algorithm}[ht]
\begin{algorithmic}
\caption{Ranking context}\label{alg:phoneme}
\State $H \gets$ best ASR hypothesis
\State $\alpha \gets$ index search threshold 
\State $r \gets$ range threshold, $s \gets$ coverage threshold
\State $L \gets []$, $S \gets []$, $R \gets []$

\State $C \gets index\_search(H, \alpha)$
\State $H^p \gets grapheme\_to\_phoneme (H)$
\For {$C_i$ in C}
    \State $C_i^p \gets grapheme\_to\_phoneme (C_i)$
    \State $LCS_i \gets LCS(H^p, C_i^p)$
    \State $L_i \gets len(LCS_i)$
    \State $S_i \gets \frac{L_i}{len(C_i^p)}$
    \State $R_i \gets$ range of $LCS_i$ in $H^p$
\EndFor

\State $L_{MAX} \gets max(L)$
\State $j \gets index (L, L_{MAX})$
\If {$S_j \geq s$ and $\frac{R_j}{len(C_j^p)} \leq r$}
    \State $candidate = C_j$
    \Else
        \State $S_{MAX} \gets max(S)$
        \State $j \gets index (S, S_{MAX})$
        \If {$S_j \geq s$ and $\frac{R_j}{len(C_j^p)} \leq r$}
            \State $candidate = C_j$
        \EndIf
\EndIf

\If {$candidate$ not assigned}
    \State \Return no correction proposed
\EndIf

\State $N \gets$ rewrite of $H$ by matching its tokens with the ones in $candidate$ covered by $LCS_j$
\State \Return $N$
\end{algorithmic}
\end{algorithm}

Algorithm \ref{alg:phoneme} shows the implementation of ranking context. We used $\alpha = 0.5$, $r = 1.5$, and $s = 0.8$. We tuned $r$ and $s$ based on several ASR errors in real user dialogues initially and a handful of false positives in real user dialogues after deployment. $\alpha$ was determined by manually inspecting several examples during post-hoc analysis.

\section{Detailed implementation of GPT augmentation}\label{augment-detail}
Algorithm \ref{alg:cap} summarizes the process of our index creation. We prepared a \textit{public} wikiHow dataset \cite{Koupaee2018wikihow}, 
which contains $230K$ instances of related tasks for augmentation, in addition to our private list of $50K$ available wikiHow tasks to broaden the range of potential correction scenarios and created a mapping between these instances ($(X, Y)$ in Algorithm \ref{alg:cap}).
We used MiniLM \cite{wang2020minilm} for obtaining embeddings, Kmeans with default scikit-learn\footnote{\url{https://scikit-learn.org/stable/modules/generated/sklearn.cluster.KMeans.html}} parameters for clustering, and GPT-3.5-turbo\footnote{\url{https://platform.openai.com/docs/models/gpt-3-5}} for generating variations. We used 20K clusters with k=8, resulting in 160K additional variations. 

\begin{algorithm}[ht]
\begin{algorithmic}
\caption{Context-augmented Index Creation}\label{alg:cap}
\State $M \gets \{\}$, $Q \gets$ query 
\State $X \gets$ public dataset, $Y \gets$ private dataset
\State $E_X \gets $ embed. (X), $E_Y \gets $ embed. (Y)
\State $\alpha \gets$ similarity threshold
    \For {$x_i,y_i$ in (X,Y)}
        \State $S \gets cosine\_sim (E_Y(y_i), E_X(x_i))$  

\If {$S > \alpha$} $M[x_i] = y_i$ \EndIf
\EndFor
\State $X' \gets \{x \in M\}, E_{X'} \gets $ Embed. (X')
\State $C_1, C_2, ... C_n \gets Centroids(Cluster(E_{X'}))$
\For {$i=1,...i=n$}
\State $C_{i1},.. C_{ik} \gets$ generate k variations
    \For {$j=1,...j=k$}
        \State $M[C_{ij}] = M[C_i]$
\EndFor
\EndFor
\end{algorithmic}
\end{algorithm}

\section{Evaluation of GPT augmentation}
\label {search-examples}
Table \ref{tab:gpt-examples} shows examples of variations generated by GPT. A manual inspection of 100 samples showed that \textbf{94\%} of the variations generated by GPT were accurate variations of the original title. In a few cases (6\%), we observed that the variations were not accurate.

\begin{table}[ht]
\centering
\begin{tabular}{p{3.1cm}|p{3.8cm}}
\hline
\textbf{Original} & \textbf{GPT variations}\\
\hline
learn quickly when reading & absorb information faster while reading\\
\hline
start a computer & 
boot up computer\\
\hline
kill mosquitoes &exterminate mosquitos\\
\hline
make cinnamon sugar & 
\textcolor{red}{bake cookies with cinnamon honey and sugar}\\
 \hline
\end{tabular}
\caption{Examples of GPT-augmented data for indexed search. Variations allow for a wider semantic search. The text in red shows possible deviations GPT may make.}
\label{tab:gpt-examples}
\end{table}

We manually evaluated the differences between our retrieval results and the default retrieval system on 32 queries and observed that in \textbf{24\%} of cases, our approach resulted in slight or significant improvement, while in 3.5\% of cases, this resulted in slightly worse performance. Before returning the search results to the user, our top result was injected as the third option in the retrieval list from the default search API (containing $10$ results), after removing any duplicates. This was done to minimize the effect of any bad search results returned by the indexed search. Examples of retrievals by our method are provided in Table \ref{tab:search-eval}.

\begin{table*}
\centering
\begin{tabular}{p{3.5cm}|p{4.5cm}|p{6cm}}
\hline
\textbf{User query} & \textbf{Context-aware retrieval} & \textbf{Default API retrieval (top 3)}\\
\hline
\multirow{4}*{clean my carpet} &\multirow{4}*{how to clean carpets} & how to spot clean carpet \\
\cline{3-3}
& &how to clean carpet without a carpet cleaner\\
\cline{3-3}
& &how to steam clean carpet \\
\cline{3-3}
\hline
\multirow{3}*{workout plans} &\multirow{3}*{how to make a workout plan} & how to cancel plans with a friend\\
\cline{3-3}

& & how to compare drug plans\\
\cline{3-3}

& & how to do a tabata workout\\
\hline
\end{tabular}
\caption{Our proposed approach retrieves more relevant results compared to the default search API}
\label{tab:search-eval}
\end{table*}

\section{Examples of failure cases}\label{examples}
Table \ref{tab:not-corrected} shows the errors in the search intent that were not corrected by our method. Table \ref{tab:fp-select} and Table \ref{tab:fp-search} demonstrate false positives because of and not because of a narrow context, respectively.

\begin{table*}[]
    \centering
    \begin{tabular}{p{4cm}|p{3.5cm}|p{3.5cm}|p{3.5cm}}
    Reason & Best ASR (wrong) & Correct transcript & Our correction \\
    \hline
    Errors in key phrases and correct transcript not in $n$-best (41\%) & how can i die my hair & how can i dye my hair & (no correction proposed) \\
    \hline
    Errors in proper nouns not in the dataset (14\%) & help me win monkey metal in blue tower defense six & help me win monkey meadow in bloons tower defense six & (no correction proposed) \\
    \hline
    Errors with prepositions (12\%) & how to get rid on ground bees & how to get rid of ground bees & (no correction proposed) \\
    \hline
    The correct transcript in $n$-best does not have a good search result (12\%) & how to stretch press muscle & how to stretch breast muscle & (no correction proposed) \\
    \hline
    The Wrong hypothesis scored high (12\%) & how do i determine correct height for a white kane & how do i determine correct height for a white cane & how do i determine correct height for a cane \\
    \end{tabular}
    \caption{Examples of ASR errors in the search intent that were not corrected by our method.}
    \label{tab:not-corrected}
\end{table*}

\begin{table*}[]
    \centering
    \begin{tabular}{p{6cm}|p{3.5cm}|p{6cm}}
    Reason & Best ASR (correct) & Our correction \\
    \hline
    A user wanted to leave our system (39\%) & go to youtube & how to get youtube (from the search results) \\
    \hline
    More specific queries or self-correction of ASR errors after searching (36\%) & turn on an alarm on a smartphone & how to test a security alarm (after our system searched only for ``alarm'')\\
    \hline
    Number in one of the ASR hypotheses after search results are presented (8\%) & let's do cycling & let's two cycling (our system selects the second option) \\
    \hline
    Chit chat (5\%) & tell me a dad joke & how to tell a dad joke (from the search results) \\
    \end{tabular}
    \caption{Examples of false positives due to a narrow context.}
    \label{tab:fp-select}
\end{table*}

\begin{table*}[]
    \centering
    \begin{tabular}{p{6cm}|p{3.5cm}|p{6cm}}
    Reason & Best ASR (correct) & Our correction \\
    \hline
    High score for an incorrect hypothesis (40\%) & how to create an app & how to create an apple \\
    \hline
    Removal of (sub) words (27\%) & how to grow grass in my backyard & how to grow grass in my yard \\
    \hline
    Bad search results for all hypotheses (24\%) & how do you take apart a macbook & how do you start a macbook \\
    \hline
    High score for an ungrammatical hypothesis (9\%) & how to do bubble braids & how to you do bubble braids \\
    \end{tabular}
    \caption{Examples of false positives when a narrow context does not exist or is not the cause.}
    \label{tab:fp-search}
\end{table*}

\end{document}